\documentclass[sigconf]{acmart}
\AtBeginDocument{%
  }

\setcopyright{acmlicensed}
\copyrightyear{2026}
\acmYear{2026}
\setcopyright{cc}
\setcctype{by}
\acmConference[WWW '26] {Proceedings of the ACM Web Conference 2026}{April 13--17, 2026}{Dubai, United Arab Emirates.}

\acmDOI{10.1145/3774904.3792089}
\acmBooktitle{Proceedings of the ACM Web Conference 2026 (WWW '26), April 13--17, 2026, Dubai, United Arab Emirates}
\acmISBN{979-8-4007-2307-0/2026/04}
\usepackage{booktabs}
\usepackage{caption}
\usepackage{multirow}
\usepackage[table,xcdraw]{xcolor}
\usepackage{pifont}
\usepackage{enumitem}
\usepackage{tcolorbox}

\begin{document}

\title[HingeMem: Boundary Guided Long-Term Memory with Query Adaptive Retrieval for Scalable Dialogues]{HingeMem: Boundary Guided Long-Term Memory with \\ Query Adaptive Retrieval for Scalable Dialogues}


\author{Yijie Zhong}
\email{zhongyj@tongji.edu.cn}
\orcid{0000-0002-2351-8799}
\affiliation{
    \institution{College of Design and Innovation, Tongji University}
    \city{Shanghai}
    \country{China}
}

\author{Yunfan Gao}
\email{gaoyunfan1602@gmail.com}
\orcid{0000-0002-7932-2752}
\affiliation{
    \institution{Shanghai Research Institute for Intelligent Autonomous Systems, Tongji University}
    \city{Shanghai}
    \country{China}
}

\author{Haofen Wang}
\authornote{Corresponding author.}
\orcid{0000-0003-3018-3824}
\email{carter.whfcarter@gmail.com}
\affiliation{
    \institution{College of Design and Innovation, Tongji University}
    \city{Shanghai}
    \country{China}
}

\renewcommand{\shortauthors}{Yijie Zhong, Yunfan Gao, and Haofen Wang}

\begin{abstract}
Long-term memory is critical for dialogue systems that support continuous, sustainable, and personalized interactions. However, existing methods rely on continuous summarization or OpenIE-based graph construction paired with fixed Top-\textit{k} retrieval, leading to limited adaptability across query categories and high computational overhead. In this paper, we propose HingeMem, a boundary-guided long-term memory that operationalizes event segmentation theory to build an interpretable indexing interface via boundary-triggered hyperedges over four elements: person, time, location, and topic. When any such element changes, HingeMem draws a boundary and writes the current segment, thereby reducing redundant operations and preserving salient context. To enable robust and efficient retrieval under diverse information needs, HingeMem introduces query-adaptive retrieval mechanisms that jointly decide (a) \textit{what to retrieve}: determine the query-conditioned routing over the element-indexed memory; (b) \textit{how much to retrieve}: control the retrieval depth based on the estimated query type. Extensive experiments across LLM scales (from 0.6B to production-tier models; \textit{e.g.}, Qwen3-0.6B to Qwen-Flash) on LOCOMO show that HingeMem achieves approximately $20\%$ relative improvement over strong baselines without query categories specification, while reducing computational cost (68\%$\downarrow$ question answering token cost compared to HippoRAG2). Beyond advancing memory modeling, HingeMem's adaptive retrieval makes it a strong fit for web applications requiring efficient and trustworthy memory over extended interactions.
\end{abstract}




\maketitle

\section{Introduction}

\begin{figure}[t]
    \centering
    \includegraphics[width=0.8\linewidth]{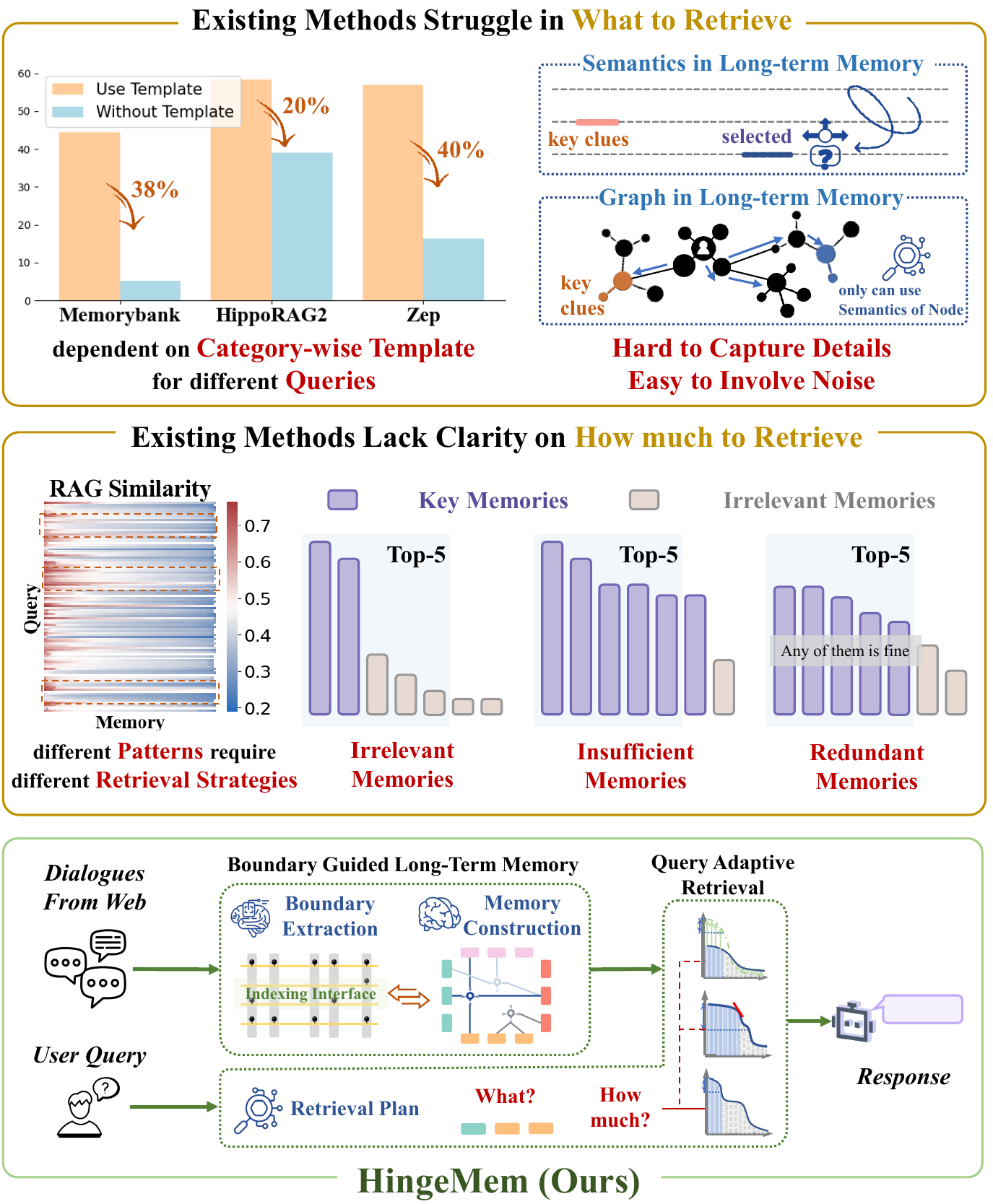}
    \caption{Existing methods face two challenges when facing various queries: they struggle to identify key clues and find it difficult to control the retrieval process. HingeMem simulates the human cortex and hippocampus to construct boundary-guided long-term memory, employing query-adaptive retrieval to generate unique retrieval plans.}
    \label{fig:intro}
\end{figure}

As Large Language Models (LLMs) advance rapidly, dialogue systems increasingly power everyday assistants and long-lived companions~\cite{DBLP:journals/corr/abs-2504-09915,DBLP:journals/corr/abs-2303-14524}. 
In human–human communication, people rely on memory to retain and leverage past information~\cite{DBLP:journals/corr/abs-2504-15965}. Intelligent Web applications built on dialogue systems similarly require robust long-term memory to support continual and personalized interactions~\cite{DBLP:conf/acl/0001YHH0LSCPLI025}. This paper focuses on advancing long-term memory modeling for dialogue systems.

Recent studies have explored long-term memory from multiple perspectives~\cite{DBLP:journals/tois/ZhangHZJCZ19,DBLP:conf/iva/SieberK10,DBLP:conf/sigdial/AsriSSZHFMS17}, including lightweight plugins~\cite{reminisc,memary}, centralized services~\cite{DBLP:journals/corr/abs-2504-19413,DBLP:journals/corr/abs-2501-13956}, and system-level architectures~\cite{DBLP:journals/corr/abs-2507-03724}. These approaches typically follow two directions: (i) continuously recording, summarizing, and updating dialogue histories as memory~\cite{DBLP:conf/emnlp/BaeKKLKJKLPS22,DBLP:journals/corr/abs-2403-04787,DBLP:conf/emnlp/SunCWHWWZY24,DBLP:journals/ijon/WangFCWTD25}, or (ii) transforming conversations into structured graph-based memory using techniques such as OpenIE~\cite{DBLP:journals/ipm/ZhongWGZWW25,zhong2025odda,DBLP:journals/corr/abs-2502-14802,DBLP:journals/corr/abs-2405-19686}. 
However, both directions often overlook details of event dynamics in the dialogue history and struggle to flexibly control retrieval across diverse queries. When faced with diverse queries without explicit category specification, responses are unstable and performance degrades substantially. Moreover, memory construction and maintenance incur high computational costs without commensurate gains~\footnote{LangMem incurs a token overhead exceeding 1000$\times$ the total tokens in all dialogues}, which limits practical deployment in Web applications.

Continuous memory writing and unconstrained OpenIE processing frequently result in incomplete detail capture and substantial overhead. Existing methods rely on semantic similarity-based retrieval over plain text or graph nodes, which hinders the precise alignment of query information and key evidence for various queries. As illustrated in~\autoref{fig:intro}, performance drops around 30\% when query categories are unspecified. Moreover, most methods employ fixed Top-\textit{k} retrieval strategies (\textit{e.g.} Memorybank~\cite{DBLP:conf/aaai/ZhongGGYW24} Top-5, Mem0~\cite{DBLP:journals/corr/abs-2504-19413} Top-30), which may introduce both redundancy and noise. The effectiveness of such fixed-\textit{k} quickly saturates or even declines for different categories of queries~\cite{DBLP:journals/corr/abs-2502-12110}. This effect becomes especially evident in real dialogue scenarios. Similarity patterns in the heatmap of vanilla RAG in~\autoref{fig:intro} indicate that fixed-\textit{k} retrieval is suboptimal. Some queries require retrieving nearly all relevant segments (\textit{e.g.} `How many meetings have we held in total?' and `What sports do I enjoy?'). Whereas others are best addressed with only a few or any one of the relevant segments (\textit{e.g.} `What was I doing at 12:00 yesterday?' and 'Have I dined with Alex before?'). Forcing the retrieval of low-relevance content or discarding high-relevance content leads to decreases in accuracy and efficiency. 

In summary, accurate and efficient long-term memory must address two complementary challenges: 1) \textbf{What to retrieve}: design a general, interpretable, and composable indexing interface that aligns the query with the right evidence elements and their combinations. 2) \textbf{How much to retrieve}: determine retrieval depth adaptively, matching the evidence size required by different queries.

Inspired by cognitive neuroscience, we propose HingeMem, a boundary-guided long-term memory with query-adaptive retrieval for dialogue systems.
Drawing on event segmentation theory, HingeMem exposes an interpretable and generalizable indexing interface over four key elements: person, time, location, and topic. 
When any of these elements changes, HingeMem writes a boundary-aligned segment as a structured hyperedge, reducing redundant operations while preserving salient details.
These hyperedges connect element-specific nodes to concise descriptions, yielding indices that support both semantic and graph-informed retrieval.
During retrieval, HingeMem generates query-adaptive retrieval plans that jointly decide what and how much to retrieve.  
The plan identifies element constraints and priorities, re-ranks candidate hyperedges, and applies adaptive stopping policies.
We categorize queries into three retrieval-oriented types: recall-priority, precision-priority, and judgment, each with tailored stopping criteria. 
Experiments across LLM scales show that HingeMem achieves superior performance with reduced computational cost, and it operates without category-specific question templates, making it adaptable to diverse scenarios.
The contributions are as follows:
\begin{itemize}[leftmargin=*, itemindent=0pt]
    \item We introduce HingeMem, a neuro-inspired, boundary-guided memory with query-adaptive retrieval. It writes memory segments at element changes, minimizing continuous writing and providing explicit, interpretable retrieval interfaces.
    \item We propose a query-adaptive retrieval mechanism that optimizes what to and how much to retrieve, overcoming the limitations of fixed-depth retrieval and reducing computational cost.
    \item Experiments show consistent gains on LOCOMO without relying on category-specific templates and show robust effectiveness across LLM scales, highlighting HingeMem's practicality for Web applications requiring efficient long-term memory.
\end{itemize}

\begin{figure*}[t]
    \centering
    \includegraphics[width=0.9\linewidth]{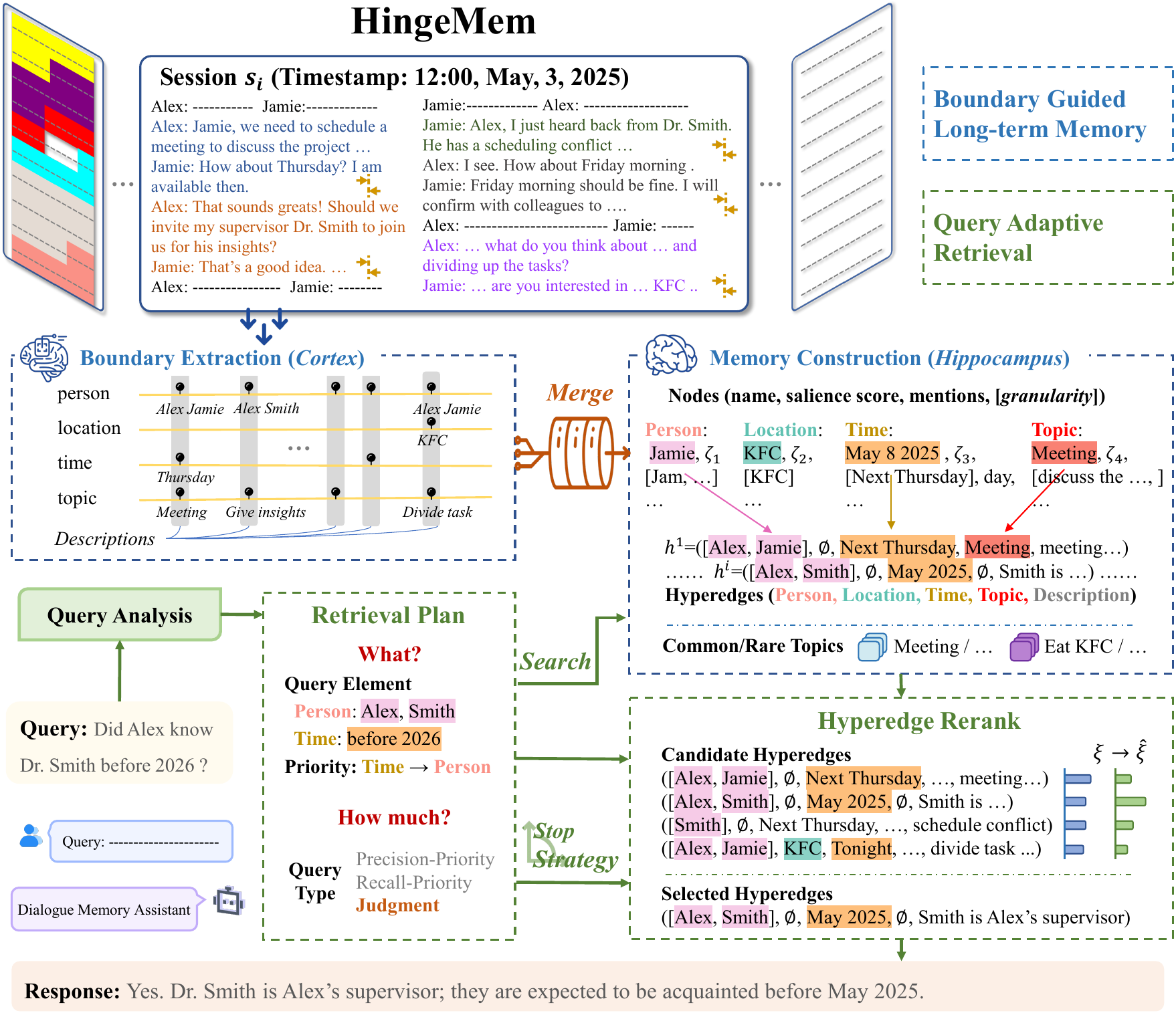}
    \caption{Overall Boundary-guided Long Term Memory construction and Query Adaptive Retrieval process of HingeMem.}
    \label{fig:pipeline}
\end{figure*}

\section{Related Work}

\subsection{Long-term Memory Mechanism}

Prior work on long-term memory for dialogue systems aims to explicitly retain and retrieve historical dialogues and user profiles~\cite{DBLP:journals/corr/abs-2504-15965}. Most methods instantiate memory as either unstructured plain text or structured graphs. 
On the text side, SCM~\cite{wang2024enhancinglargelanguagemodel} introduces a Self-Controlled Memory framework with a memory stream and controller. TiM~\cite{DBLP:journals/corr/abs-2311-08719} distills the `thoughts' in raw dialogues to support recalling and post-thinking. A line of work~\cite{DBLP:journals/corr/abs-2308-08239,DBLP:journals/ijon/WangFCWTD25} employs different summarization strategies to construct unstructured memories. However, purely textual memory often struggles to preserve fine-grained details and lacks explicit indices for targeted retrieval. On the graph side, following GraphRAG~\cite{DBLP:journals/corr/abs-2404-16130}, several studies exploit graphs to build structured memories. EMG-RAG~\cite{DBLP:conf/emnlp/WangLJTS24} maintains an editable personal knowledge graph to support insertion and deletion of memories. Several efforts leverage neuro‑symbolic techniques to build query‑specific retrieval paths~\cite{DBLP:conf/acl/GeRCSBSZ25,DBLP:conf/naacl/OngKGCKJHLY25}, but these approaches predominantly target temporal questions and are computationally expensive. While graphs provide better structural information than plain text, unconstrained extraction may introduce noise, making query alignment non-trivial for various queries.

Beyond engineering choices, researchers have drawn on insights from cognitive neuroscience~\cite{brain}. Memorybank~\cite{DBLP:conf/aaai/ZhongGGYW24} leverages the Ebbinghaus forgetting curve to guide memory updates. A representative example is HippoRAG~\cite{DBLP:conf/nips/GutierrezS0Y024,DBLP:journals/corr/abs-2502-14802}, which operationalizes hippocampal indexing theory by treating the LLM as an artificial `neocortex' and maintaining a schema-free OpenIE knowledge graph as a `hippocamal-like index'. 
This architecture is practical, yet it relies on a single unified index for the entire events with fixed Top-\textit{k} retrieval. In contrast, we draw on event segmentation theory and hippocampal–cortical interactions. We encode memory at event boundaries and construct hyperedges.
Versus text-based systems, our boundary-triggered hyperedges preserve fine-grained details and expose explicit indices. 
Versus OpenIE-centric graph memories, our element-indexed hyperedges improve interpretability and query alignment. 
Versus fixed-depth retrieval, we employ query-adaptive stopping tied to retrieval-oriented query types.

\subsection{Long-term Memory Evaluation}

For many years, researchers have developed datasets through various methods, including crowd-sourcing, web data collection, and generation~\cite{DBLP:journals/corr/abs-2505-00675,DBLP:journals/corr/abs-2504-15965}, to facilitate the evaluation and technical iteration of long-term memory for dialogue systems.
MSC~\cite{DBLP:conf/acl/XuSW22} and Conversation Chronicles~\cite{DBLP:conf/emnlp/JangBK23} serve as a foundational resource for early research. However, they are constrained by the conversation length and the correlation between dialogues.
DuLeMon~\cite{DBLP:conf/acl/XuGWNW0W22} focuses on long-term persona memory dialogues in Chinese contexts but is limited by its lack of scenario diversity.
TimelineQA~\cite{DBLP:conf/acl/TanD0MSYH23} concentrates exclusively on time series reasoning, neglecting other memory elements.
Memorybank~\cite{DBLP:conf/aaai/ZhongGGYW24} and PerLTQA~\cite{DBLP:journals/corr/abs-2402-16288} push forward comprehensive evaluations. Memorybank is limited in scale and lacks annotations, and PerLTQA's scenarios are relatively uniform
Recent benchmarks broaden the perspective:
MADial-Bench~\cite{DBLP:conf/naacl/HeZWWHZ25} evaluates how long-term memory supports user emotions, while LAMP~\cite{DBLP:conf/acl/SalemiMBZ24} emphasizes whether responses can mimic the user's personality and behavior.
SHARE~\cite{DBLP:conf/acl/KimPC25} and HiCUPID~\cite{DBLP:conf/acl/MokKPY25} explore long-term dialogues but face singular memory source and narrow question category constraints.

LongMemEval~\cite{DBLP:conf/iclr/WuWYZCY25} and LOCOMO~\cite{DBLP:conf/acl/MaharanaLTBBF24} are most related to our work. LongMemEval establishes a benchmark for evaluating long-term interactive memory. However, each dialogue contains only a single question, making a comprehensive assessment inefficient. 
In contrast, LOCOMO focuses on ultra-long dialogue memory with a data design closely aligned with realistic multi-turn scenarios, offering advantages in memory span and question diversity. Accordingly, we adopt LOCOMO and follow its category-wise F1 score and an additional LLM-as-a-Judge setting for comparability.

\section{HingeMem}

\subsection{Overview}

\autoref{fig:pipeline} presents our proposed neuro-inspired HingeMem, comprising two key components: (a) Boundary-Guided Long-Term Memory, which leverages node indices and hyperedges; and (b) Query-Adaptive Retrieval, which features retrieval planning, hyperedge reranking, and adaptive stopping. A simulated \textit{cortex} performs dialogue boundary extraction within each session, extracts key elements from segments, and organizes them into a hyperedge. A simulated \textit{hippocampus} then consolidates the resulting hyperedges and element-indexed nodes into long-term memory. During retrieval, HingeMem produces a retrieval plan for each query, explicitly specifying both \textit{what} to retrieve and \textit{how much} to retrieve. 

\subsection{Boundary Guided Long-Term Memory}

Cognitive neuroscience indicates that the brain does not encode information uniformly. Instead, it privileges event boundaries~\cite{baldassano2017discovering,geerligs2022partially}. At such transitions, activity in the prefrontal cortex (PFC) intensifies, hippocampal activation increases, and interactions between the hippocampus and PFC, collectively facilitating memory formation~\cite{reagh2020aging,zheng2022neurons}. Converging evidence further shows that indexing at event offsets or boundaries is beneficial, whereas mid-event indexing can be counterproductive~\cite{franklin2020structured,lu2022neural}.

Event Segmentation Theory (EST)~\cite{zacks2007event} holds that segmentation is automatic~\cite{zacks2001perceiving}, operates at multiple scales~\cite{biederman1987recognition}, and is triggered by salient changes in time, space, objects or characters, goals, or causality~\cite{biederman1987recognition}. Such boundaries partition continuous streams, reducing representational load and supporting memory. Critically, boundary-aligned information is more easily recalled, and temporal order is more reliably preserved. Recent fMRI work further underscores the importance of “encoding at boundaries,” showing that hippocampal–cortical interactions at these points promote persistent and detailed memory representations~\cite{barnett2024hippocampal}. Therefore, we design boundary-centric encoding in HingeMem by coupling a cortex-like boundary extractor with a hippocampus-like memory constructor.

\subsubsection{Dialogue Boundary Extraction}

Inspired by Event Segmentation Theory, we identify boundaries in dialogues based on changes in key elements. In dialogue systems, these changes are primarily reflected in the addition or alteration of \textit{person, time, location}, and \textit{topic}. Given $l$ sessions denoted as $S=\{s_1,\cdots,s_l\}$, natural dialogue boundaries are formed during session transitions (\textit{e.g.} from $s_i$ to $s_{i+1}$). In session $s_i$, we simulate the cortex to extract boundaries. The content between two boundaries forms what we refer to as boundary-guided memory. 

To facilitate more intuitive indexing and semantic information, we establish nodes based on the various elements (\textit{i.e.} person, time, location, and topic) within this memory and create an indexing interface. Building upon this semantic information, we further connect nodes of different elements within the same segment to form a hyperedge, representing the boundary-guided memory. Specifically, we construct a boundary extraction prompt $P_{BE}$ and invoke LLMs ($\phi$) to obtain the memory $B_i$, which contains the list of element nodes $N_i$ and hyperedges $H_i$ for $s_i$, as shown below:
\begin{equation}
    \phi(s_i | P_{BE}) \rightarrow B_i=(N_i, H_i).
\end{equation}
The list $N_i$ consists of person nodes $P_i$, time nodes $T_i$, location nodes $L_i$, and topic nodes $C_i$. We can define any node $n$ as follows:
\begin{equation}
    \textbf{node}: n = (\text{name}, \text{mentions}, \text{[optional] granularity}),
\end{equation}
where \textit{name} serves as a unique identifier in $s_i$ and \textit{mentions} refers to the list of mentions that appears in $s_i$. When $n\in T_i$, additional granularity regarding time is extracted to present its finest division, thus preventing the model from generating incorrect time due to hallucination. 
The hyperedge $h^j$ in $H_i$ is mainly composed of a subset ($\tilde{P}^j, \tilde{T}^j, \tilde{L}^j, \tilde{C}^j$) of nodes from all the elements associated with the corresponding segment ($P_i, T_i, L_i, C_i$), as shown below:
\begin{equation} 
    \textbf{hyperedge}: h^j = (\tilde{P}^j, \tilde{T}^j, \tilde{L}^j, \tilde{C}^j, d^j, r^j),
\end{equation}
We utilize the description $d_i$ to retain semantics from several turns within each hyperedge. Furthermore, we constrain the model to analyze the reasons $r_i$ for segmenting the boundaries, thereby enhancing the accuracy of extracting key elements that undergo change. Specifically, $r_i$ is selected from \{\textit{change person, change time, change location, topic shift}, and \textit{explicit marker}\}.

\subsubsection{Memory Construction}

Leveraging the hippocampus's role in the continuous integration of new information into long-term memory, we utilize hippocampal-cortical interactions to guide the sustained encoding of boundary memories $B_i$ and enhance the enduring retention of details in $s_i$. This process ($M\rightleftharpoons B_i,\ i=1\cdots l$) enables establishing boundary-guided long-term memory $M$.

The core of this process lies in the effective integration of new and existing boundary memories. For each node, we merge relevant mentions based on unique identifiers. During this process, timestamps are converted into ISO 8601 format according to the time granularity. For instance, the term \textit{yesterday} is transformed into the `\%Y-\%m-\%d' format based on the session time or a specific time in the session. Additionally, we compute the salience score of each node to reflect its importance within the overall dialogue. The salience score takes into account the following three dimensions: \\

\noindent \textbf{Frequency}: This measures the number of times the node appears in long-term memory, thus determining its significance.

\noindent \textbf{Centrality}: This analyzes the degree of the node within the hypergraph structure, as more central nodes are typically more important.

\noindent \textbf{Diversity}: This assesses the co-occurrence of the node with others, reflecting how common it is across different contexts. \\

\noindent For hyperedges, we determine whether to merge them by calculating the field-aware Jaccard Score ($J$) between pairs of hyperedges ($h^i$ and $h^j$). First, we compute a node set for each hyperedge $N_{(h^i)}$ and then compute the score for any two hyperedges as follows: 
\begin{equation}
    J(h^i,h^j) = |N_{(h^i)}\cap N_{(h^j)}| / |N_{(h^i)}\cup N_{(h^j)}|.
\end{equation}
We proceed to merge hyperedges recursively until no pair of hyperedges has a Jaccard score exceeding 0.8. This strategy not only enhances the storage efficiency of long-term memory by avoiding redundancy but also ensures the accuracy and relevance of the information in boundary-guided memories.

We further analyze all hyperedges and utilize the theme clustering prompt $P_{TC}$ to identify common topics $C_{common}$ that frequently appear in long-term memory, as well as rare topics $C_{rare}$ that are mentioned less often. This information is crucial for developing subsequent retrieval strategies. To accommodate different categories of user queries, we should deploy varied recall strategies based on the frequency of mentions. For example, we should assess the quantity of recalls for common topics while emphasizing specific key pieces of memory for rare topics.
The boundary-guided long-term memory can finally be represented as:
\begin{equation}
    M = \{N, H, C_{common}, C_{rare}\}.
\end{equation}

Thus, we have established a boundary-guided long-term memory and provided four explicit element node indices. By incorporating the structured information from hyperedges, the system effectively delivers both semantic and graph-related information.

\subsection{Query Adaptive Retrieval}

\subsubsection{Retrieval Plan Generation}

To address the diverse categories of queries, the dialogue system needs to analyze these queries and provides targeted retrieval plans that clarify `\textit{what}' the system needs to retrieve and `\textit{how much}' historical content is required. Specifically, we employ a query analysis prompt $P_{Q}$ to process the queries, resulting in retrieval plans that encompass the following key information: the predicted type of query, the element and name to be queried, and the element priority.  

Regarding query types, to ensure the applicability of the system, we do not directly predict the categories of queries delineated in specific datasets (such as LOCOMO). Instead, we categorize queries into three types based on the retrieval process: Recall Priority, Precision Priority, and Judgment.
Recall priority signifies that it is essential to retrieve as many hyperedges in $M$ related to the query as possible for an effective response. Precision priority, on the other hand, indicates that the most relevant individual or a few hyperedges are required to answer the question, rendering other relevant memories unnecessary in this context. Judgment query only requires identifying any a few of relevant pieces of memory to answer, as excessive retrieval may lead to redundancy.

Regarding query elements, the retrieval plan specifies the relevant interfaces that need to be queried for the query (\textit{i.e.} elements person, time, location, and topic). Additionally, $P_{Q}$ also asks to extract the \textit{names} that provide information under these interfaces.

Regarding element priority, we believe that the order of priority for query interfaces is crucial for the adaptive retrieval of each query. For example, in the query `When did Caroline have lunch at KFC?', the priority of time is significantly higher than person and location. Therefore, the retrieval plan includes the ranking of different elements ($p$) to enhance the retrieval process.

From a structural perspective, based on the query elements and their associated names, HingeMem identifies the corresponding nodes ($\hat{N}$). This enables system to obtain hyperedges that encompass these nodes. From a semantic perspective, HingeMem matches the query with the similarity of the embeddings of the hyperedges' descriptions. Consequently, HingeMem derives a set of candidate hyperedges ($\hat{H}$) from the boundary-guided long-term memory $M$, with each hyperedge's initial score ($\xi$) being its similarity value.

\subsubsection{Hyperedge Rerank}

A straightforward selection from the similarity between query ($q$) and hyperedges ($\hat{H}$) can lead to suboptimal outcomes, particularly in scenarios with diverse query types and varying relevance of retrieved memories. To enhance the retrieval process, we propose a reranking mechanism by incorporating the relationships among hyperedges within long-term memory.
This mechanism prioritizes hyperedges based on the salience values of the involved nodes, ensuring that more pertinent memories are emphasized. 
Furthermore, we propose to add a penalty term to prevent less frequent memories from being overshadowed in the long-term memory.
This penalty term considers the proximity of the hyperedges to identified rare and common topics, allowing for a nuanced distinction between widely referenced and less frequent yet relevant information.

The computation of impact ($\Omega_S$) of salience values involves a weighted sum of the elements based on their priority $p$ as obtained in the retrieval plan. The element scores are obtained by averaging the salience values of all nodes in the corresponding node list (such as $\tilde{P}^i$).
For the topic penalty term ($\Omega_T$), we calculate the distance from the current query to rare terms and subtract the distance to common terms. 
The definitions of rare terms and common terms are derived from the respective rare and common topics. To mitigate the issue of overlapping topics between the $C_{common}$ and $C_{rare}$, we first calculate the feature subspaces for both rare and common topics. Subsequently, we remove any overlapping subspaces and assess the cross-similarity to generate weights accordingly. After computing the similarity between the query and each topic in $C_{common}$ and $C_{rare}$, we finally apply a weighted softmax aggregation to derive the rare and common terms for the specific query.
The updated score $\hat\xi^i$ for each hyperedge $\hat{h}^i$ with its initial score $\xi^i$ according to query $q$ is computed as follows:
\begin{equation}
    \hat{\xi^i} = \xi^i + \Omega_S(\hat{N}_{(\hat{h}^i)}|p^i) + \Omega_T(q|C_{common},C_{rare}).
\end{equation}
This strategy to reranking ensures that the final selection of hyperedges is not only more relevant to the current query but also presents a balanced view of existing personal knowledge in the long-term memory by considering both common and rare details.

\begin{figure}[t]
    \centering
    \includegraphics[width=\linewidth]{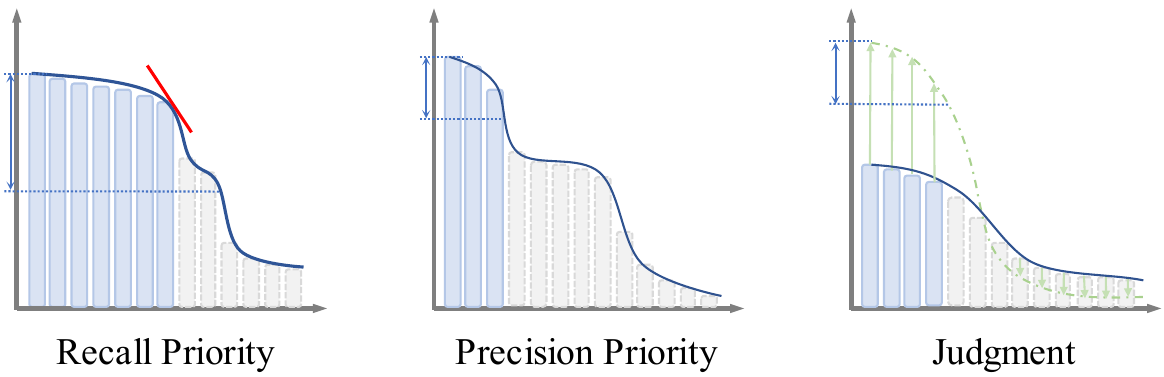}
    \caption{Diagram of adaptive stop for different query types.}
    \label{fig:adapt}
\end{figure}

\begin{table*}[t]
\centering
\caption{
Results on the LOCOMO dataset across five question categories. We report F1 scores ($F_1$), BLEU-1 scores ($B_1$), and LLM-as-a-Judge scores ($J$). The higher values indicate better performances. $\dagger$ denotes scores reported by the A-Mem paper. `Cat.' indicates whether to apply the different QA format for each category. Best results are shown in \textbf{Bold}.
}
\resizebox{0.8\linewidth}{!}{
\begin{tabular}{@{}lclcccccccccccccccccc@{}}
\toprule
\multicolumn{1}{c}{} &  &  & \multicolumn{2}{c}{\textbf{Single-Hop}} &  & \multicolumn{2}{c}{\textbf{Multi-Hop}} &  & \multicolumn{2}{c}{\textbf{Temporal}} &  & \multicolumn{2}{c}{\textbf{Open-Domain}} &  & \multicolumn{2}{c}{\textbf{Adversarial}} &  & \multicolumn{3}{c}{\textbf{Overall}} \\ \cmidrule(lr){4-5} \cmidrule(lr){7-8} \cmidrule(lr){10-11} \cmidrule(lr){13-14} \cmidrule(lr){16-17} \cmidrule(l){19-21} 
\multicolumn{1}{c}{\multirow{-2}{*}{\textbf{Method}}} & \multirow{-2}{*}{\textbf{Cat.}} &  & \textbf{$F_1$} & \textbf{$J$} &  & \textbf{$F_1$} & \textbf{$J$} &  & \textbf{$F_1$} & \textbf{$J$} &  & \textbf{$F_1$} & \textbf{$J$} &  & \textbf{$F_1$} & \textbf{$J$} & \multirow{-2}{*}{} & \textbf{$F_1$} & \textbf{$J$} & \textbf{$B_1$} \\ \midrule
\multicolumn{1}{l|}{LOCOMO} & \multicolumn{1}{c|}{\ding{51}} &  & 12.7 & 16.5 &  & 19.7 & 20.9 &  & 10.4 & 11.5 &  & 20.1 & 30.2 &  & 66.8 & 90.1 & \multicolumn{1}{c|}{} & 25.8 & 33.5 & 0.132 \\
\multicolumn{1}{l|}{RAG (Top-5)} & \multicolumn{1}{c|}{\ding{51}} &  & 32.4 & 47.4 &  & 26.9 & 38.6 &  & 22.5 & 18.0 &  & 21.4 & 35.4 &  & \textbf{90.6} & \textbf{90.3} & \multicolumn{1}{c|}{} & 42.6 & 50.5 & 0.293 \\
\multicolumn{1}{l|}{RAG (Top-10)} & \multicolumn{1}{c|}{\ding{51}} &  & 36.4 & 50.7 &  & 29.6 & 38.6 &  & 27.7 & 22.7 &  & 21.2 & 37.5 &  & 86.8 & 86.5 & \multicolumn{1}{c|}{} & 44.6 & 51.9 & 0.306 \\
\multicolumn{1}{l|}{RAG (Top-20)} & \multicolumn{1}{c|}{\ding{51}} &  & 37.7 & 52.5 &  & 34.9 & 48.9 &  & 30.0 & 22.7 &  & 21.9 & 41.6 &  & 83.2 & 82.9 & \multicolumn{1}{c|}{} & 45.5 & 53.5 & 0.315 \\ \midrule
\multicolumn{1}{l|}{ReadAgent$^\dagger$} & \multicolumn{1}{c|}{\ding{51}} &  & 6.6 & - &  & 2.5 & - &  & 5.3 & - &  & 10.1 & - &  & 5.4 & - & \multicolumn{1}{c|}{} & 5.7 & - & 0.109 \\
\multicolumn{1}{l|}{MemGPT$^\dagger$} & \multicolumn{1}{c|}{\ding{51}} &  & 10.4 & - &  & 4.2 & - &  & 13.4 & - &  & 9.5 & - &  & 31.5 & - & \multicolumn{1}{c|}{} & 14.6 & - & 0.125 \\
\multicolumn{1}{l|}{A-Mem$^\dagger$} & \multicolumn{1}{c|}{\ding{51}} &  & 18.2 & - &  & 24.3 & - &  & 16.4 & - &  & 23.6 & - &  & 46.0 & - & \multicolumn{1}{c|}{} & 25.2 & - & 0.208 \\ \midrule
\multicolumn{1}{l|}{Memorybank} & \multicolumn{1}{c|}{\ding{55}} &  & 6.8 & 63.0 &  & 9.1 & 33.6 &  & 4.6 & 34.8 &  & 5.6 & 35.2 &  & 0.0 & 59.1 & \multicolumn{1}{c|}{} & 5.2 & 52.0 & 0.031 \\
\multicolumn{1}{l|}{{\color[HTML]{9B9B9B} \textit{+ Cat. format}}} & \multicolumn{1}{c|}{\ding{51}} &  & 39.9 & 59.2 &  & 23.3 & 28.7 &  & 23.3 & 31.7 &  & 20.9 & 44.7 &  & 86.3 & 86.3 & \multicolumn{1}{c|}{} & 44.4 & 55.8 & 0.303 \\ \midrule
\multicolumn{1}{l|}{LangMem} & \multicolumn{1}{c|}{\ding{55}} &  & 54.7 & 70.1 &  & 47.4 & 58.5 &  & 48.6 & 51.0 &  & 27.8 & 38.3 &  & 17.3 & 81.1 & \multicolumn{1}{c|}{} & 42.9 & 62.5 & 0.307 \\
\multicolumn{1}{l|}{{\color[HTML]{9B9B9B} \textit{+ Cat. format}}} & \multicolumn{1}{c|}{\ding{51}} &  & 54.1 & 70.1 &  & 45.6 & 58.5 &  & 46.8 & 51.7 &  & 21.5 & 44.1 &  & 87.5 & 87.0 & \multicolumn{1}{c|}{} & 57.6 & 64.3 & 0.392 \\ \midrule
\multicolumn{1}{l|}{HippoRAG2} & \multicolumn{1}{c|}{\ding{55}} &  & 54.4 & 78.5 &  & 35.4 & 52.8 &  & 55.5 & 61.0 &  & 23.4 & 35.2 &  & 4.3 & 69.5 & \multicolumn{1}{c|}{} & 39.1 & 68.5 & 0.289 \\
\multicolumn{1}{l|}{{\color[HTML]{9B9B9B} \textit{+ Cat. format}}} & \multicolumn{1}{c|}{\ding{51}} &  & 59.2 & 75.5 &  & 38.0 & 46.4 &  & 44.9 & 66.6 &  & 21.2 & 35.4 &  & 87.7 & 87.2 & \multicolumn{1}{c|}{} & 58.4 & 70.6 & 0.396 \\ \midrule
\multicolumn{1}{l|}{Mem0} & \multicolumn{1}{c|}{\ding{55}} &  & 45.1 & 56.7 &  & 42.7 & 48.2 &  & 49.7 & 50.1 &  & 27.7 & \textbf{47.2} &  & 6.5 & 56.6 & \multicolumn{1}{c|}{} & 36.0 & 53.7 & 0.254 \\
\multicolumn{1}{l|}{{\color[HTML]{9B9B9B} \textit{+ Cat. format}}} & \multicolumn{1}{c|}{\ding{51}} &  & 44.0 & 59.0 &  & 38.6 & 47.8 &  & 45.6 & 47.6 &  & 20.8 & 42.0 &  & 84.3 & 72.7 & \multicolumn{1}{c|}{} & 51.4 & 59.6 & 0.351 \\ \midrule
\multicolumn{1}{l|}{Mem0$_g$} & \multicolumn{1}{c|}{\ding{55}} &  & 45.3 & 55.1 &  & 40.4 & 48.5 &  & 47.9 & 52.0 &  & 28.4 & 41.0 &  & 6.7 & 54.1 & \multicolumn{1}{c|}{} & 35.5 & 51.7 & 0.258 \\
\multicolumn{1}{l|}{{\color[HTML]{9B9B9B} \textit{+ Cat. format}}} & \multicolumn{1}{c|}{\ding{51}} &  & 43.4 & 62.0 &  & 38.3 & 45.7 &  & 44.4 & 48.5 &  & 23.4 & 44.1 &  & 82.7 & 68.5 & \multicolumn{1}{c|}{} & 50.7 & 59.7 & 0.348 \\ \midrule
\multicolumn{1}{l|}{Zep} & \multicolumn{1}{c|}{\ding{55}} &  & 20.6 & 70.0 &  & 22.0 & 52.4 &  & 21.7 & 52.6 &  & 6.6 & 40.4 &  & 2.9 & 54.1 & \multicolumn{1}{c|}{} & 16.3 & 59.6 & 0.098 \\
\multicolumn{1}{l|}{{\color[HTML]{9B9B9B} \textit{+ Cat. format}}} & \multicolumn{1}{c|}{\ding{51}} &  & 59.4 & 76.7 &  & 41.1 & 52.8 &  & 50.0 & \textbf{66.9} &  & 22.4 & 40.0 &  & 74.7 & 74.4 & \multicolumn{1}{c|}{} & 56.9 & 69.4 & \textbf{0.406} \\ \midrule
\multicolumn{1}{l|}{\textbf{HingeMem}} & \multicolumn{1}{c|}{\ding{55}} &  & \textbf{61.1} & \textbf{78.8} &  & \textbf{53.6} & \textbf{62.8} &  & \textbf{57.4} & \textbf{66.9} &  & \textbf{30.7} & 46.4 &  & 87.4 & 87.8 & \multicolumn{1}{c|}{} & \textbf{63.9} & \textbf{75.1} & 0.404 \\ \bottomrule
\end{tabular}}
\label{tab:total}
\end{table*}

\subsubsection{Adaptive Stop}

Unlike existing methods, we do not simply select the Top-k hyperedges as context. Instead, we adopt an adaptive approach to determine the appropriate number of memory contents based on the query type. This strategy not only reduces overall unnecessary token costs but also mitigates the risk of introducing excessive distracting information, which could lead to incorrect answers. For details on when to stop selecting relevant hyperedges, refer to~\autoref{fig:adapt}. Here, we first sort the updated score $\hat{\xi}$.

\noindent For \textbf{Recall Priority} query, we detect inflection points in the score changes. Specifically, we select hyperedges $\hat{h}^i$ before the index $\min\{i:\hat{\xi}^{i+1}-\hat{\xi}^i < \lambda_{knee} \wedge \hat{\xi}^i > \max(\hat\xi)/2\}$~\footnote{In practice, we set $\lambda_{knee}$ to 0.1.}. 
This enables us to focus on a sufficient number of relevant memories while avoiding the noise that may arise from lower-quality hyperedges.

\noindent For \textbf{Precision Priority} query, we prioritize the most confident data segments. Specifically, we dynamically scale the acceptable confidence to select hyperedges where the score ($\hat\xi^i$) exceeds 80\% of the maximum score $\max(\hat\xi)$. This ensures that the retrieved hyperedges are not only relevant but also reliable, thereby enhancing the overall response accuracy.

\noindent For \textbf{Judgment} query, we aim to select data that has a significant impact on the query. We initially apply a softmax function to scale all candidate hyperedges. In this way, even if many hyperedges that fulfill the query have relatively low scores, we can still select any one of them after rescaling. Subsequently, we choose hyperedges with softmax scores greater than 80\% of the maximum softmax score. This strategy emphasizes the hyperedges that are likely to provide the most valuable information for making judgments.

Ultimately, the selected hyperedges serve as the context for generating the final answer to the current query. Notably, HingeMem does not require the use of a specific question template for different categories of queries, such as those provided by LOCOMO~\footnote{For example, Line 243-259 in \textit{task\_eval/gpt\_utils.py} in its official repository.}.

\section{Experiment}

\subsection{Experimental Settings}

\begin{table}[t]
\centering
\caption{Statistics of the LOCOMO dataset.}
\resizebox{\linewidth}{!}{
\begin{tabular}{@{}lc|lc@{}}
\toprule
\textbf{Conversation Statistics} & \textbf{Counts} & \textbf{Question Statistics} & \textbf{Counts} \\ \midrule
Total Conversations & 10 & Single-hop Questions & 841 \\
Avg. Sessions in a conversation & 27.2 & Multi-hop Questions & 282 \\
Avg. Turns in a conversation & 294.1 & Temporal Questions & 321 \\
Avg. Tokens in a conversation & 15965.8 & Open-domain Questions & 96 \\
Avg. Tokens in a session & 587.0 & Adversarial Questions & 446 \\
Avg. Tokens in a Turn & 27.1 & Total & 1986 \\ \bottomrule
\end{tabular}}
\label{tab:dataset}
\end{table}

\subsubsection{Datasets}

We evaluate the long-term conversational memory in dialogue systems using the LOCOMO~\cite{DBLP:conf/acl/MaharanaLTBBF24} dataset, which contains substantially longer dialogues than previous datasets, such as MSC~\cite{DBLP:conf/acl/XuSW22} and Conversation Chronicles~\cite{DBLP:conf/emnlp/JangBK23}. As summarized in~\autoref{tab:dataset}, LOCOMO comprises multi-session conversations averaging 15K tokens and spanning up to 27 sessions
~\footnote{Data is sourced from the LOCOMO Github repository and differs from the version reported in the original paper. The authors provide a more representative subset to enable efficient validation for the community. In contrast, LongMemEval~\cite{DBLP:conf/iclr/WuWYZCY25} contains only 500 questions, each tied to a single dialogue. This makes evaluation inefficient and environmentally unfriendly. Therefore, we primarily use LOCOMO in this paper.}, 
making it well-suited for assessing systems’ ability to construct long-term memory and maintain cross-session consistency over extended interactions.
Each dialogue involves two participants discussing daily experiences or past events over a few months. Following each conversation, the dataset provides $\sim$200 questions with corresponding ground-truth answers, covering five categories: single-hop, multi-hop, temporal, open-domain, and adversarial.

\subsubsection{Metrics}

Following the previous research in conversational AI~\cite{metric,10738994}, we report the lexical metrics F1 Score ($F_1$) and BLEU-1 ($B_1$). While answers with substantial lexical overlap may still contain critical factual errors. To mitigate this limitation, we also employ an LLM-as-a-Judge ($J$) as a complementary evaluation metric. The judge model analyzes the question, ground-truth answer, and the system's answer, delivering a more nuanced evaluation that better aligns with human judgment. To ensure a fair comparison, we follow the LOCOMO evaluation protocol: computing $F_1$ scores separately for each question category with different rules and then aggregating to obtain the overall $F_1$ scores. For the LLM-as-a-Judge, we apply the same prompt template as Mem0~\cite{DBLP:journals/corr/abs-2504-19413}.

\subsubsection{Baselines}

We compare the proposed HingeMem against two types of baselines distinguished by their memory structures:
(1) \textbf{Only Semantics}: LOCOMO~\cite{DBLP:conf/acl/MaharanaLTBBF24}, Retrieval-Augmented Generation (RAG)~\cite{DBLP:conf/acl/MaharanaLTBBF24}, ReadAgent~\cite{DBLP:conf/icml/LeeCFCF24}, MemGPT~\cite{DBLP:journals/corr/abs-2310-08560}, A-Mem~\cite{DBLP:journals/corr/abs-2502-12110}, Memorybank~\cite{DBLP:conf/aaai/ZhongGGYW24}, and Mem0~\cite{DBLP:journals/corr/abs-2504-19413};
(2) \textbf{Semantics and Graphs}: LangMem, HippoRAG2~\cite{DBLP:conf/nips/GutierrezS0Y024,DBLP:journals/corr/abs-2502-14802}, Mem0 with graph memory (Mem$_g$)~\cite{DBLP:journals/corr/abs-2504-19413}, and Zep~\cite{DBLP:journals/corr/abs-2501-13956}.
For a fair comparison, we conduct experiments using their open-source implementation under identical experimental settings. Unless otherwise specified, all experiments are conducted with \texttt{GPT-4o} via the official structured output API. We use \texttt{text-embedding-3-small} to achieve all the text embeddings. \\


\begin{figure}[t]
    \centering
    \includegraphics[width=\linewidth]{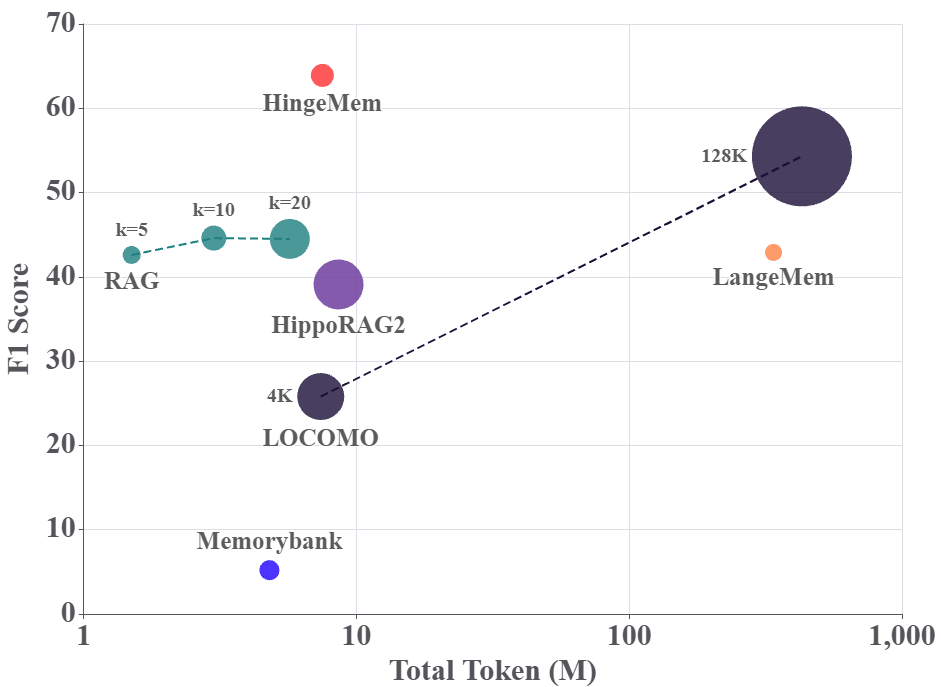}
    \caption{A comparative analysis of efficiency across different methods. For LOCOMO, we employ two configurations with context lengths of 4K and 128K. For RAG, we establish several parameter settings ranging from Top-5 to Top-20. The horizontal axis represents the total number of tokens consumed during the entire memory construction (if applicable) and the question-answering process. The circle size corresponds to the total tokens expended for all the users' queries.}
    \label{fig:token}
\end{figure}

\subsection{Main Results}

\subsubsection{Superiority.}

\autoref{tab:total} demonstrates that HingeMem achieves optimal performance across various categories of queries. Notably, it shows an over 5\% improvement in overall metrics and an impressive enhancement exceeding 10\% on multi-hop questions. This significant progress is primarily attributed to our proposed boundary-guided memory, which effectively captures more memory details from chaotic dialogues. Although RAG (Top-5) achieves the best results on adversarial questions, HingeMem remains competitive in performance. RAG provides limited information by selecting only the top five pieces of content. Furthermore, it often retrieves irrelevant information in extensive dialogues, making it hard to fall into hallucinations when confronted with LLM's powerful contextual comprehension ability and the input of short contexts.

Additionally, HingeMem, with the proposed adaptive retrieval, excels without the need to know the question type. In contrast, existing methods experience substantial performance declines when not informed of the category of questions. Memorybank and Zep show the most significant drops, ranging from 30\% to 40\%. 
Overall, HingeMem reveals its superiority across multiple critical metrics, confirming its efficacy in dealing with complex conversations for practical long-term memory.

\subsubsection{Efficiency.}

We conduct experiments to assess the impact of different methods and settings on efficiency. For Mem0 and Zep, we are unable to obtain an accurate token cost as we use their official services. Therefore, we exclude them in this comparison. The results are presented in~\autoref{fig:token}. While the memory construction process can be completed offline, the computation cost during the inference phase in response to user requests is of greater significance. We report the total token consumption during the experiments as well as the token cost associated with the question-answering phase.

The results indicate that increasing the context size (to 128K) allows for the transmission of more personalized information to the dialogue system, thereby significantly enhancing performance. However, this improvement comes with an unacceptable computational cost, reaching up to 0.4 billion tokens. Vanilla RAG alleviates the high computational demand by retrieving relevant content from historical dialogue to be incorporated into the context, while maintaining moderate performance. Nevertheless, it fails to effectively build long-term personal memory, as it requires searching through the entire history, which increases the inference overhead. Furthermore, vanilla RAG complicates the maintenance of historical data, making it more challenging to quickly and accurately locate the information needed by the user.

As~\autoref{tab:total} illustrates that constructing long-term memory contributes to improved system performance.~\autoref{fig:token} further reveals that leveraging long-term memory can significantly reduce computational costs during the question-answering phase. Notably, HippoRAG2 stands as an exception, sustaining computational costs similar to those of RAG and LOCOMO. This is attributed to its use of unconstrained OpenIE to organize complex graph memories, necessitating additional attention during retrieval. It can also be observed that the memory construction of LangMem appears to be excessively inefficient compared to other methods. In contrast, our HingeMem demonstrates superior performance with total token and inference token costs comparable to existing approaches.

\begin{table}[t]
\centering
\caption{Ablation study. TM and BM represent the common Textual Memory and our Boundary Guided Memory. \textcircled{2} only uses the hyperedges' description for retrieval. NI means Node Indexing, which utilizes the structural information in hyperedges. HR and AS represent the proposed Hyperedge Rerank and Adaptive Stop.}
\resizebox{0.8\linewidth}{!}{
\begin{tabular}{@{}c|l|c|ccccc@{}}
\toprule
\multicolumn{1}{l|}{\textbf{\# Num}} & \textbf{Method} & \multicolumn{1}{l|}{\textbf{All}} & \multicolumn{5}{c}{\textbf{Each Category}} \\ \midrule
\textcircled{1} & RAG+TM & 44.6 & 36.4 & 29.6 & 27.7 & 21.2 & 86.8 \\
\textcircled{2} & RAG+BM & 57.4 & 54.5 & 39.3 & 50.1 & 24.3 & 86.8 \\
\textcircled{3} & RAG+BM+NI & 58.1 & 56.1 & 43.6 & 51.2 & 27.7 & 82.7 \\
\textcircled{4} & \textcircled{3}+HR & 61.2 & 59.5 & 47.4 & 52.9 & 28.2 & 86.1 \\
\textcircled{5} & \textcircled{3}+HR+AS & 63.9 & 61.1 & 53.6 & 57.4 & 30.7 & 87.4 \\ \bottomrule
\end{tabular}}
\label{tab:ab}
\end{table}

\begin{figure}[t]
    \centering
    \includegraphics[width=\linewidth]{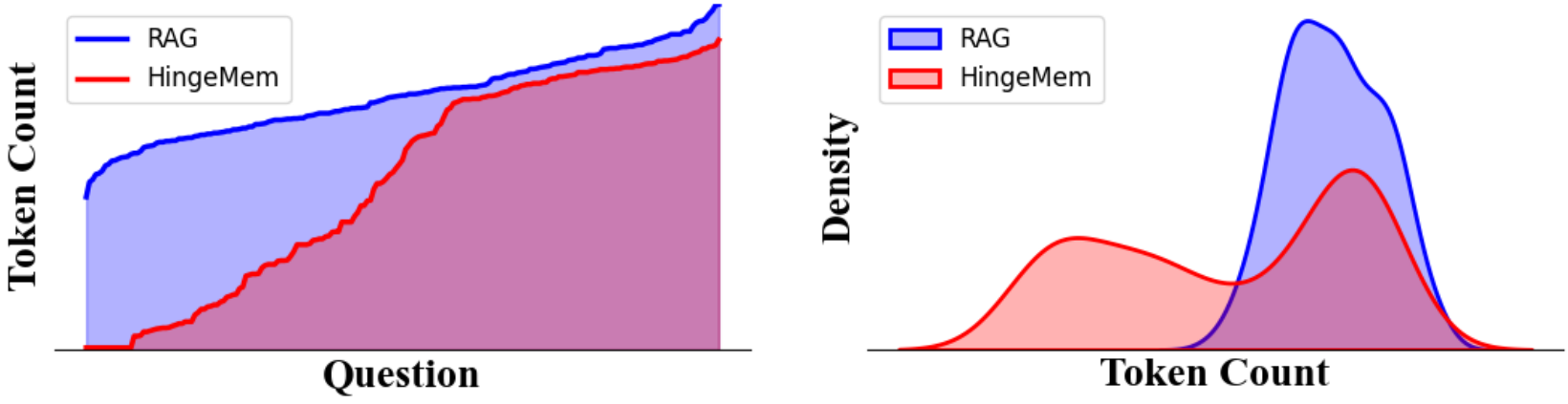}
    \caption{Analysis of the token count and its distribution of retrieval results for the question in sample 'conv-26' in LOCOMO, comparing RAG and HingeMem. HingeMem adaptively sets the retrieval capacity for different categories of questions, thereby reducing unnecessary noise and ultimately enhancing performance and efficiency.}
    \label{fig:context}
\end{figure}

\subsection{Ablation Study}

\subsubsection{Effectiveness of Boundary Guided Memory.}

We conduct experiments using vanilla RAG on both plain text-based memory and the proposed boundary guided memory to investigate the superiority of the boundary-guided memory. The results are illustrated in~\autoref{tab:ab}. 
Comparing \textcircled{1} and \textcircled{2}, we find that boundary-guided memory significantly enhances overall performance, resulting in an improvement of over 10\%. This benefits from its ability to capture more detailed segments in the dialogues. However, because it does not intervene in the retrieval process, only using boundary-guided memory cannot reduce the noise memory present in the final selected context, resulting in no improvement for adversarial questions. Building on semantic retrieval, we further introduce node indexing based on information related to elements in the query to incorporate additional structural information. The results in \textcircled{3} indicate that the integration of semantics and structure leads to further performance improvements. Particularly for multi-hop questions, we observe a near 4\% improvement.

\subsubsection{Effectiveness of Query Adaptive Retrieval.}

We conduct a detailed evaluation of the role of the proposed adaptive retrieval from both performance and efficiency perspectives. 
As shown in~\autoref{tab:ab}, \textcircled{4} indicates a slight improvement following the introduction of hyperedge rerank. This enhancement arises from the strategy's consideration of the relationship between query and long-term memory, as well as the influences among different pieces of memory within the long-term memory. A comparison between \textcircled{4} and \textcircled{5} demonstrates that the dialogue system achieves optimal performance across various categories of questions by utilizing our adaptive stop strategy.

Query Adaptive Retrieval not only ensures the response accuracy but also reduces unnecessary token costs. We compare the context lengths retrieved by RAG and HingeMem for the same set of questions, as shown in~\autoref{fig:context}. The context length obtained by Top-K fluctuates within a certain range, making it vulnerable to noise interference and the loss of related information. In HingeMem, adaptive retrieval enables the specification of targeted retrieval plans for each query, thereby reducing overall token cost while ensuring that the retrieval results are informative and free of redundancy.

\subsubsection{Robustness of Different Model Scales.}

\begin{figure}[t]
    \centering
    \includegraphics[width=\linewidth]{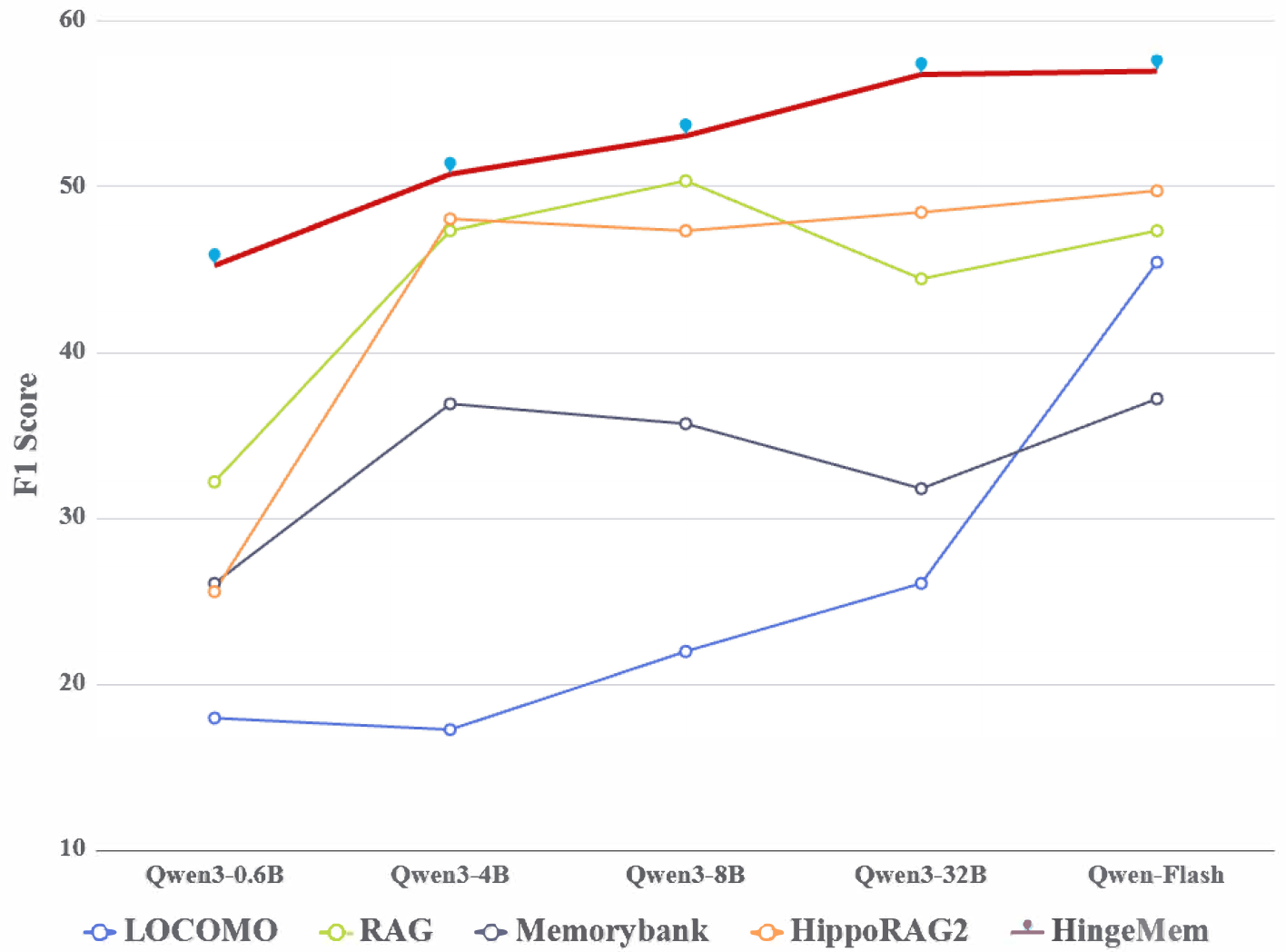}
    \caption{The performance comparison of different model scales in the Qwen-Series models. HingeMem delivers consistent and optimal results from the 0.6B model to the flagship model. This demonstrates the adaptability of our approach across various potential application scenarios.}
    \label{fig:qwen}
\end{figure}

We conduct evaluations on the advanced Qwen3-Series~\cite{DBLP:journals/corr/abs-2505-09388} model of different scales to validate the generalization ability of existing methods. The results in~\autoref{fig:qwen} show that HingeMem maintains stable and superior performance across all scales, from the 0.6B model to the flagship model, demonstrating strong adaptability to changes in model size. Specifically, as the model scale increases, the performance curves exhibit a consistently rising trend, whereas the other baselines display fluctuations or declines at certain scales. This suggests that existing methods are relatively dependent on powerful base LLMs. Notably, in small-scale models, the vanilla RAG setup already achieves performance that surpasses other methods with explicitly constructed memory. These findings indicate that our proposed HingeMem generalizes robustly across computing powers and remains effective not only on web servers but also on resource-constrained mobile edge devices, enabling adaptive inference from cloud to on-device deployment with broad application potential.

\section{Conclusion}

In this work, we introduce HingeMem, inspired by hippocampal-cortical interactions and event segmentation theory. HingeMem extracts the boundary memory by locating the changes in key elements of events: person, time, location, and topic. Then it constructs the structured hyperedges from these elements and updates the boundary guided long-term memory. By incorporating adaptive retrieval strategies, HingeMem addresses the critical challenges of efficient memory retrieval for queries of different categories. Extensive experiments demonstrate that HingeMem outperforms existing state-of-the-art methods, offering a 5\% improvement on LOCOMO while maintaining scalability across various scales of LLMs. HingeMem not only provides a practical solution to the limitations of existing memory mechanisms but also opens new avenues for research in memory design and efficient knowledge retrieval for conversational AI and intelligent Web applications.

\section*{Acknowledgements}

This work was supported by the National Natural Science Foundation of China (U23B2057), and Shanghai Pilot Program for Basic Research (22TQ1400300).

\bibliographystyle{ACM-Reference-Format}
\balance
\bibliography{sample-base}

\appendix

\section{More Implementation Details}
\label{sec:app-a}

To enable fair comparison, we integrate all baselines into our project based on the following open-source codes. Then we use the same code for evaluation. Below, we focus on the most critical component: the prompt design. 

\begin{itemize}[leftmargin=*, itemindent=0pt]
    \item LOCOMO and RAG: \textit{https://github.com/snap-research/LoCoMo}
    \item Memorybank: \\ \textit{https://github.com/zhongwanjun/MemoryBank-SiliconFriend}
    \item HippoRAG2: \textit{https://github.com/OSU-NLP-Group/HippoRAG}
    \item LangMem and Mem0: \textit{https://mem0.ai/research} and \\
    \textit{https://github.com/mem0ai/mem0/tree/main/evaluation}
    \item Zep: \textit{https://github.com/getzep/zep-papers} and \\ \textit{https://www.getzep.com/}
\end{itemize}

\subsection{HingeMem}

We present the Boundary Extract Prompt, Topic Clustering Prompt, and Query Analysis Prompt in our proposed HingeMem for reference. Here, we preserve the output schemas while omitting several additional principles in the prompt.

\begin{tcolorbox}[title={Prompt: Boundary Extraction},opacityframe=0.7,fontupper=\small]
You are a "conversation segmentation and element extractor". Please perform the following tasks on the input complete conversation session (including datetime, turn id, and speaker name) and output only valid JSON.

\# Task Objectives

1. Identify event or memory boundaries: When there is an obvious change in **person/time/location/topic**, or a new **person/time/location/topic** appears, or when explicit transition words appear, start a new event; otherwise, merge it into the current memory data.

2. Extract relations and events in a unified form.

Fields: persons[], times[], locations[], topics[], description, boundary{reasons[], start turn, end turn}

For relations: 
    (optional)
    **Person - Person**: ... ;
    **Person - Time**: ... ;
    **Person - Location**: ... 

For events:
    Fill in the corresponding fields according to the events involved and summarize the corresponding topics.

...

\# Output Specifications:

...

\# Output Schema

\{
    "persons": [
        \{"canonical\_name": "<string>", "role\_tags": ["<string>", ...], "mentions": [\{"turn": <int>, "mention": "<string>"\}, ...]\}, 
        ...
    ], \\
    "times": [
        \{"timestamp": "<ISO8601> (do not provide null, estimate based on the relative date if possible)", "granularity": "second | minute | hour | day | week | month | year | approx", "mentions": [\{"turn": <int>, "mention": "<string>"\}, ...]\},
        ...
    ], \\
    "locations": [
        \{"name": "<string>", "mentions": [\{"turn": <int>, "mention": "<string>"\}, ...]\},
        ...
    ], \\
    "topics": [
        \{"label": "<string>", "mentions": [\{"turn": <int>, "mention": "<string>"\}, ...]\},
        ...
    ], \\
    "boundary\_memories": [
        \{
            "person\_list": ["<canonical\_name mentioned above>", ...],
            "time\_list": ["<timestamp mentioned above>", ...],
            "location\_list": ["<name mentioned above>", ...],
            "topic\_list": ["<label mentioned above>", ...],
            "description": "<string>",
            "boundary": \{
                "reasons": ["change\_time | change\_place | change\_person | topic\_shift | explicit\_marker"], 
            \}
        \},
        ... 
    ],
    ... 
\}

***

\{session data\}
\end{tcolorbox}

\begin{tcolorbox}[title={Prompt: Topic Clustering},opacityframe=0.7,fontupper=\small]
Task: From INPUT\_TOPICS (a list of historical event topics), select only from the input and return two lists.

Definitions: \\
- Common/Stable: widely documented, recurring across periods \\
- Rare/Uncommon: niche/localized, sparsely documented.

Rules: \\
- Pick 3–5 topics for each list; ... \\
- Concise noun phrases only; no dates or explanations. \\
- Do not invent; deduplicate; preserve original phrasing/case. \\
- Output valid JSON only (no extra text).

Return exactly this JSON shape: \\
\{"common\_topics": ["..."], "rare\_topics": ["..."]\}

INPUT\_TOPICS: 
\{topics\}
\end{tcolorbox}

\begin{tcolorbox}[title={Prompt: Query Analysis},opacityframe=0.7,fontupper=\small]
[Role] \\
You are the "Query Analyzer". Please read the input query and output a structured search plan. The four elements you can use are: person, location, topic, time.

[Analyze Principles] \\
- Question Type Definition (Choose One): \\
    - Recall-First: Questions that require coverage / enumeration / counting / timeline summarization. For example: 
        - ... \\
    - Precision-First: Questions that seek the most relevant/single best piece of evidence. For example: 
        - ... \\
    - Judgment-First: Yes / No / Existential decisions, or a balance of evidence. For example:
        - ... \\
- If the person/location/topic/time is not clear, leave the corresponding fields blank and do not make them up. \\
- "priority" only includes elements that exist in "constraints"

[JSON Schema] \\
\{ \\ 
    "query\_type": "recall | precision | judgement", \\
    "constraints": \{ \\
        (optional) "person": ["str", ...], \\
        (optional) "location": ["str", ...], \\
        (optional) "topic": ["str", ...], \\
        (optional) "time": [\{ \\
            "timestamp": "(ISO Time Format)", \\
            "granularity": "year | month | day | hour | minute | approx" \\
        \}, ...], \\
    \}, \\
    "priority": [(sorted keys in constraints)]  // List only the factors that you think are discriminatory, in order of priority \\
\}

[Query] \\
{{query}}

[Output JSON]
\end{tcolorbox}

\subsection{LOCOMO Template}

Below, we present the different templates used in LOCOMO for various question categories. For example, prompts for temporal questions require LLMs to return an explicit `Date'. An inspection of existing methods’ open-source code shows that they employ these category-specific prompts, which limits their practical applicability. In contrast, HingeMem adaptively handles various questions.

\begin{tcolorbox}[title={Prompt: Template for Temporal Questions},opacityframe=0.7,fontupper=\small]
Based on the above context, write an answer in the form of a short phrase for the following question. Answer with exact words from the context whenever possible.

Question: \{Question\} Use DATE of CONVERSATION to answer with an approximate date. Short answer:
\end{tcolorbox}

\begin{tcolorbox}[title={Prompt: Template for Adversarial Questions},opacityframe=0.7,fontupper=\small]
Based on the above context, answer the following question.

Question: \{Question\} Select the correct answer: (a) \{Adversarial Answer a\} (b) \{Adversarial Answer b\}. Short answer:
\end{tcolorbox}

\begin{tcolorbox}[title={Prompt: Template for Other Questions},opacityframe=0.7,fontupper=\small]
Based on the above context, write an answer in the form of a short phrase for the following question. Answer with exact words from the context whenever possible.

Question: \{Question\} Short answer:
\end{tcolorbox}

\section{Details of Constructed Long-Term Memory}

\autoref{tab:memory} presents the counts of element nodes and hyperedges in the long-term memory constructed by HingeMem on LOCOMO. The results show that HingeMem effectively captures fine-grained conversational details while avoiding the substantial redundancy and noise introduced by the continuous memory-writing strategies used in prior methods.

\begin{table}[h]
\centering
\caption{Statistics of the constructed long-term memory.}
\resizebox{0.8\linewidth}{!}{
\begin{tabular}{l|c}
\hline
\textbf{Memory Statistics} & \textbf{Counts} \\ \hline
Avg. Persons in a conversation & 12.8 \\
Avg. Times in a conversation & 59.6 \\
Avg. Locations in a conversation & 34.6 \\
Avg. Topics in a conversation & 81.5 \\
Avg. Hyperedges in a conversation & 103.2 \\ \hline
\end{tabular}}
\label{tab:memory}
\end{table}

\section{Parameter Analysis}

\begin{figure}[h]
    \centering
    \includegraphics[width=0.8\linewidth]{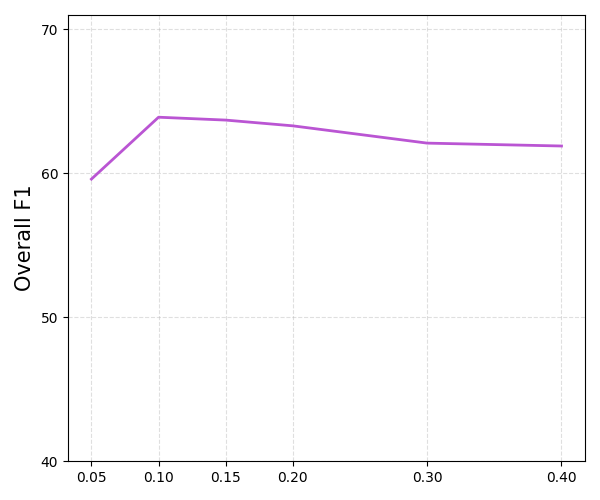}
    \caption{Performance of different values for $\lambda_{knee}$}
    \label{fig:knee}
\end{figure}

\subsection{$\lambda_{knee}$ for Recall-Priority Query}

\autoref{fig:knee} shows the effect of the hyperparameter $\lambda_{knee}$ on performance. 
If the inflection point threshold is set too small, the process halts prematurely, and valuable cues are lost.
A moderate relaxation of this threshold yields the best performance.
As $\lambda_{knee}$ further increases, performance degrades but quickly reaches a floor. 
This is because (i) memory scores often do not exhibit abrupt, large-magnitude changes, rendering larger $\lambda_{knee}$ has no practical effect, and (ii) the adaptive stopping mechanism for recall priority queries is also constrained by a maximum-score ratio, which further attenuates sensitivity to this hyperparameter.

\subsection{Scale for Precision-Priority Query}

\begin{figure}[h]
    \centering
    \includegraphics[width=0.8\linewidth]{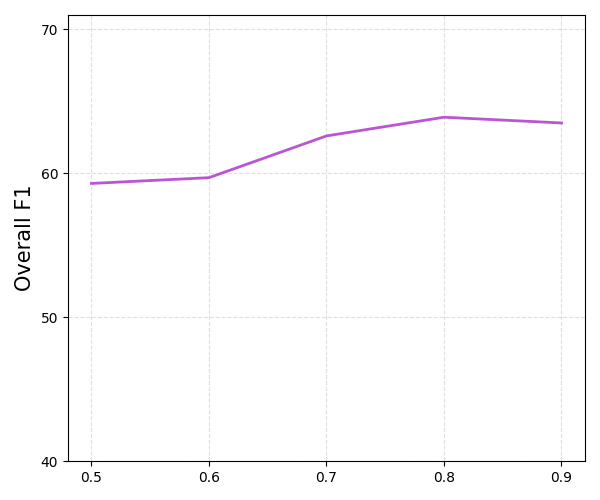}
    \caption{Impact of the scale for precision-priority query.}
    \label{fig:pre-scale}
\end{figure}

In practice, we scale the confidence to select hyperedges where the score exceeds 80\% of the maximum score for precision-priority queries.~\autoref{fig:pre-scale} shows the effect of the scale on performance. A small scale tends to admit numerous seemingly relevant yet inessential cues. To obtain more accurate and non-redundant cues, a larger scale is preferable. However, excessively large values (\textit{i.e.} 0.9) do not yield additional performance gains and may narrow the scope of applicability. Accordingly, we set the scale to 0.8.

\subsection{Scale for Judgment Query}

\begin{figure}[h]
    \centering
    \includegraphics[width=0.8\linewidth]{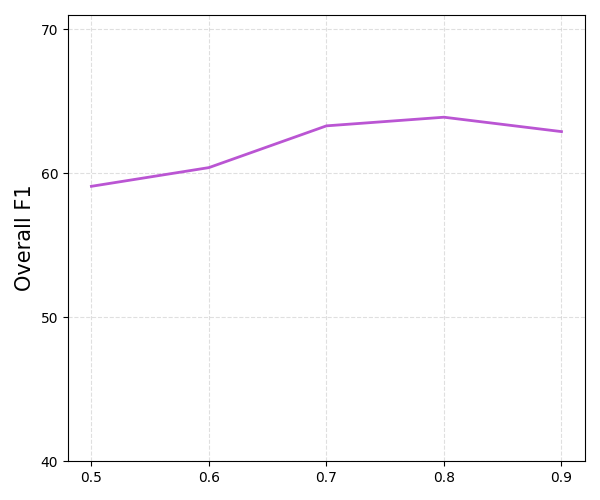}
    \caption{Impact of the scale for judgment query.}
    \label{fig:judgement}
\end{figure}

\autoref{fig:judgement} illustrates the effect of the scale applied to the relative maximum softmax score on judgment queries, exhibiting a trend similar to that observed for precision-priority queries. A small scale weakens the constraint, admitting spurious cues and degrading decision accuracy. Conversely, for judgment queries, an overly large scale has a more pronounced adverse impact on the final decision. Accordingly, we adopt a balanced setting of 0.8.

\end{document}